\documentclass[sigconf]{acmart}

\usepackage{amsmath,amssymb,amsfonts}
\usepackage{algorithmic}
\usepackage{graphicx}
\usepackage{textcomp}
\usepackage{xcolor}
\usepackage{dsfont}
\usepackage{tabularx}
\usepackage{multirow}
\usepackage{tcolorbox}

\newtcolorbox{rqanswer}{%
    colback=gray!30,
    colframe=gray!45,
}

\AtBeginDocument{%
  \providecommand\BibTeX{{%
    \normalfont B\kern-0.5em{\scshape i\kern-0.25em b}\kern-0.8em\TeX}}}

\setcopyright{acmlicensed}
\copyrightyear{2020}
\acmConference[ASE '20]{35th IEEE/ACM International Conference on Automated Software Engineering}{September21--25, 2020}{Virtual Event, Australia}
\acmYear{2020}
\acmDOI{10.1145/3324884.3416559}

\acmBooktitle{35th IEEE/ACM International Conference on Automated Software Engineering (ASE '20), September 21--25, 2020, Virtual Event, Australia}
\acmPrice{15.00}
\acmISBN{978-1-4503-6768-4/20/09}

\begin{document}

\title{Hybrid Deep Neural Networks to Infer State Models of Black-Box Systems}

\author{Mohammad Jafar Mashhadi}
\affiliation{%
  \institution{University of Calgary}
  \city{Calgary}
  \state{Canada}
}
\email{mohammadjafar.mashha@ucalgary.ca}

\author{Hadi Hemmati}
\affiliation{%
  \institution{University of Calgary}
  \city{Calgary}
  \state{Canada}
}
\email{hadi.hemmati@ucalgary.ca}


\begin{abstract}
Inferring behavior model of a running software system is quite useful for several automated software engineering tasks, such as program comprehension, anomaly detection, and testing. Most existing dynamic model inference techniques are white-box, i.e., they require source code to be instrumented to get run-time traces. However, in many systems, instrumenting the entire source code is not possible (e.g., when using black-box third-party libraries) or might be very costly. 
Unfortunately, most black-box techniques that detect states over time are either univariate, or make assumptions on the data distribution, or have limited power for learning over a long period of past behavior. To overcome the above issues, in this paper, we propose a hybrid deep neural network that accepts as input a set of time series, one per input/output signal of the system, and applies a set of convolutional and recurrent layers to learn the non-linear correlations between signals and the patterns, over time. We have applied our approach on a real UAV auto-pilot solution from our industry partner with half a million lines of C code. We ran 888 random recent system-level test cases and inferred states, over time. Our comparison with several traditional time series change point detection techniques showed that our approach improves their performance by up to 102\%, in terms of finding state change points, measured by F1 score. We also showed that our state classification algorithm provides on average 90.45\% F1 score, which improves traditional classification algorithms by up to 17\%.
\end{abstract}

\maketitle

\section{Introduction}
\label{sec:intro}
Automated specification mining or model inference \cite{lo2011mining} is the process of automatically reverse engineering a model of an existing software system. Behavioral models (e.g., state machines) are typically inferred from a running system by abstracting the execution traces. The inferred models are useful artifacts in many use cases where the actual behavior (abstracted as the inferred model) of the system is needed for analysis, such as debugging \cite{hybriddebugging, shang2013assisting, jafar2019interactive}, testing \cite{Walkinshaw2018TestingBlackBox, ModelBasedTesting, Papadopoulos2015, dallmeier2011automatically}, anomalous behavior detection \cite{valdes2000adaptive}, and requirements engineering \cite{damas2005generating}. 

Inferring a behavior model of a system in a black-box manner is particularly interesting. In many real-world applications, the large-scale system is built by integrating many off-the-shelf libraries that are only available as binaries (no source code access). Thus, from a system's point of view, knowing the exact behavior of the system including all the interactions between black-box units are needed for most run-time analysis. 

Most current behavioral model inference techniques are dynamic analysis methods (usually are more accurate than static analysis for run-time behavior inference) that require source code instrumentation to collect execution traces \cite{lo2011mining}. These methods are usually helpful in unit-level analysis where the instrumentation is not expensive and access to the code is allowed for the unit under study. However, in the system-level, thorough instrumentation is more expensive (not limited to one unit) and might not be even possible for some units (black-box libraries). Therefore, for use cases such as system-level anomaly detection, testing, and debugging a black-box behavior model inference that works on readily available input/outputs of the system is crucial. 


In this paper, we propose a dynamic analysis method to detect the internal state and the state changes in a black-box software system using deep learning. We collected the numerical values of the inputs and outputs of the system, in regular time intervals to create a multivariate time-series. A hybrid deep learning model (including convolution and recurrent layers) was then trained on these time-series to predict the state of the system at each point in time. The deep learning model automatically performs feature extraction making it way more effective and flexible compared to traditional methods. In addition, we do not make any assumption about statistical properties of the data which makes it applicable to a wide range of subjects.   

We applied and evaluated this method on an auto-pilot software (AutoPilot) used in an Unmanned Aerial Vehicle (UAV) system developed by our industry partner
\begin{anonsuppress}
Winnipeg based Micropilot Inc. Micropilot is the world-leader in professional UAV auto-pilot which develops both hardware and software for 1000+ clients (including NASA, Raytheon, and Northrop Grumman) in 85+ countries during the past 20+ years. 
\end{anonsuppress}
We evaluated the method from two perspectives: how well the model can detect the point in time when a state change happens? (RQ1: Change Point Detection (CPD)), and how accurately it can predict which state the system is in, during the execution? (RQ2: State Classification). {In addition, in RQ3, we explored some simpler variations of our method (non-hybrid variations) to roughly see how each part contributes to the overall model performance.}

Comparing our approach with state-of-the-art alternatives, the results show that our approach performs better in both change point and state detection. We observed 88.00\% to 102.20\% improvement in the F1 score of our CPD, compared to traditional CPD techniques. In addition, we saw a 7.35\% to 16.83\% improvement in the F1 score of our state detection, compared to traditional classification algorithms on a sliding window, over the data.

The contributions of this paper can be summarised as:
\begin{itemize}
    \item Introducing the first (to the best of our knowledge) deep learning architecture to infer behavior models from black-box software systems.
    \item Empirically evaluating the model and achieving very high accuracy compared to baselines using a real-world and large-scale case study on a UAV auto-pilot system developed by our industry partner.
\end{itemize}

Note that we have made all our source code, models, and execution scripts available online\footnote{https://github.com/sea-lab/hybrid-net}, however, due to confidentiality, we can not make our dataset public. 

The rest of this paper is organized as follows: In section~\ref{sec:motivation} we explain further how and in which contexts this research can be beneficial. Then in section~\ref{sec:background} we briefly explain some background material this work is based on. In section~\ref{sec:approach}, we explain how our proposed model is designed. The way it was evaluated and the results are explained in section~\ref{sec:experiment}. Finally, related work reviewed in section~\ref{sec:related_work}, and some final remarks about the future work are made in section~\ref{sec:summary}.
\section{Motivation} \label{sec:motivation}
Black-box components are ubiquitous in software development. Reusing high-quality black-box units generally offers a better overall system quality and a higher productivity \cite{edwards2001framework}.  The black-box units can be as small as a reusable library, or as large as a framework (such as .NET before going open-source), or a complete piece of software such as a remotely hosted web service. There are also scenarios where the unit's source code is not black-box in general, but not accessible to a specific team that wishes to perform the dynamic analysis. 

One particular interesting use case of system-level black-box analysis is inferring run-time state model of a control software systems, where inputs/outputs are signals to/from the system. These inputs/outputs are typically multivariate time series, which are already logged in such systems (no overhead for instrumentation). The goal is automatically detecting the high-level system state and its changes, over time.

As discussed in section~\ref{sec:intro}, we partnered with an auto-pilot manufacturer and performed this study on their auto-pilot software (called AutoPilot in this paper). The goal was to determine the state of AutoPilot from its input/output signals, over time. In this scenario, the inputs are the sensor readings going into AutoPilot and the outputs are command signals sent to controller motors of the aircraft, showing AutoPilot's reaction to each input at each state. A state in this example is the high-level stage of a flight and a state change happens when the current input values in the current state trigger a constraint in the implementation that changes the way the output signals are generated. 

In this example, the training set will consist of input and output values recorded during one execution of the system, as a multivariate time series, along with state ids (as labels) per time stamp. One execution of the AutoPilot will be the whole flight process that may go through a ``take-off'' until a successful ``landing''. Depending on the flight plan, AutoPilot goes through states such as ``acceleration'', ``take-off'', ``climbing'', ``turning'', ``descending'', etc. 



During a flight, the AutoPilot monitors changes in the input values and makes adjustments to its outputs in order to hold some invariants (predefined rules). 
For example, if AutoPilot is in the ``hold altitude'' mode, it monitors the altimeter's readings and when it goes out of the acceptable range, proportionate adjustments to the throttle or the nose pitch will be made to get it back to the desired altitude. This is basically how a typical feedback loop controller, such as PID or its variations work \cite{feedbacksystemsBook}.
When AutoPilot's state changes from ``hold altitude'' to ``descend to X ft'' state, the set of invariants that AutoPilot is trying to hold are changed. It means its reactions to variations in inputs will be different. In this example, a decreasing altimeter reading will not trigger an increase in the throttle anymore.

Looking at the time series, a domain expert can identify what the state of AutoPilot is, at each point in time; the labeling process. Now the goal is to automate this task on a test set (in practice, future flights), assuming a training set is labeled by the experts (they only need to identify the state change time stamps, during a flight). 

This problem can be tackled in two ways. The first solution is to identify the time stamp that the state change happens (i.e., Change Point Detection: RQ1); The more advanced solution is to predict the exact state per time stamp (i.e., State Classification: RQ2). The classic CPD techniques on time series \cite{Truong2018ChangePointSurvey} are mainly applicable on univariate data or put assumptions on the input/output distributions, thus not applicable in our case with multivariate inputs and no assumptions or knowledge about the states' distribution. The classic state classification techniques in time series are also weak in that 
they fail to balance between considering long-term relations or acting locally. The ones that use a sliding window, for example, do not have a long-term memory. The ones that act on the whole data on the other hand are too coarse-grained and inaccurate for this task.

Therefore, the motivation for this study is to provide a black-box technique that can be applicable on both CPD and state classification problems, and overcome the limitations of the existing techniques, in terms of capturing the non-linear correlation between multivariate inputs and outputs as well as learning patterns over a long period of time. Our proposal, which will be explained in detail in section~\ref{sec:approach}, leverages the power of a deep neural network (DNN) with two types of layers that are particularly useful for this problem: a) convolutional layers which discover latent features from the data effectively through parameter sharing and b) recurrent layers that play a significant role in problems dealing with time series as they can learn long-term dependencies and seasonalities in the data. 

Though our motivational example, as well as our case study, are from the UAV auto-pilot domain, our proposed method can be adapted to be applied to similar black-box control software systems in domains such as IoT, intelligent video surveillance, and self-driving cars.
\section{Background} \label{sec:background}
Unlike the numerous techniques in the literature for behavior model inference \cite{lang1998results, walkinshaw2016inferring, Lo2007Mining, dallmeier2006mining} which abstract a set of execution traces into states, our approach requires consuming a multivariate time-series and detect the state changes across time and predict the exact state labels. Thus, in this section, we briefly explain the two main sets of relevant existing techniques for ``Change Point Detection'' and ``State Prediction'' in time-series that can serve as background for our approach.  

\subsection{Change Point Detection}
A fundamental tool in time-series data analysis is Change Point Detection (CPD). It refers to the task of finding points of abrupt change in the underlying statistical model or its parameters that could be a result of a state transition \cite{aminikhanghahi2017survey}.
There are plenty of CPD algorithms; many of which perform effectively on a subset of CPD problems with some assumptions. The assumptions can be of various types. For example, one may assume the time series has only one input variable (univariate) \cite{fryzlewicz2014wild}, there is only one changing point \cite{bai1998testing}, or the number of change points is known beforehand \cite{lavielle2005using}, or they might assume some statistical properties on the data \cite{chen2011parametric}. These are limiting factors, since many of these assumptions do not necessarily hold in our case. CPD techniques are categorized into two main groups: a) online methods that process the data in real-time and b) offline methods that start processing the data after receiving all the values \cite{Truong2018ChangePointSurvey}. Since our model inference use case of CPD can afford waiting to collect all historical training data, we only considered offline techniques. 

In general, CPD algorithms consist of two major components: a) the search method and b) the cost function \cite{Truong2018ChangePointSurvey}.
Search methods are either exact or approximate. For instance, Pelt is the most efficient exact search method in the CPD literature, which uses pruning \cite{killick2012optimal}. Approximate methods include window-based \cite{basseville1993detection}, bottom-up \cite{keogh2001online}, binary segmentation \cite{scott1974cluster}, and more. In the window-based segmentation a sliding window is rolled over the data and then sum of costs of left and right half-windows is subtracted from the cost of the whole window. When the difference gets significantly high it means that the discrepancy between left and right half of the window is high and therefore a change point probably lies right in the middle of the window. In the bottom-up method, the input signal is split into multiple smaller parts, then using a similarity measure adjacent segments are merged until no more merges are feasible. The binary segmentation method finds one change point and splits the input into two parts around that point and then recursively applies the same method on each part.

The cost functions are also quite various, from simply subtracting each point from the mean to much more complex metrics, such as auto-regressive cost functions \cite{angelosante2012group}, and kernel-based cost functions. Kernel-based costs can have a wide variety, since the kernel function can be almost arbitrary, however a handful of them such as linear and Gaussian kernels are among the most popular ones \cite{Truong2018ChangePointSurvey}.

In the context of our paper, we need a CPD method with no assumption on data distribution, number of change points, etc. In addition, our CPD method should work on multivariate data, and be able to capture non-linear relations between signals. It also needs to be resilient to time lags between an input signal change and its effect on the output signal (and the systems state). There is no traditional CPD algorithms that covers all these requirements.
Therefore, we propose a novel CPD techniques that is based on Hybrid DNNs and compare it with several existing CPD techniques as our baselines, which are explained in details in section~\ref{sec:experiment}.

\subsection{Convolutional and Recurrent Neural Networks}
In both our problems (CPD and state classification), we can see that the changes in signals are more informative than their absolute values. Therefore, applying a derivation operation (or more generally a gradient) seems like necessary, at some point in the processing. Farid and Simoncelli listed some discrete derivation kernels in their study \cite{Farid2004}, but to have a more generalized and more flexible notion of discrete derivatives, convolutions seems like a better choice to apply. 
Nowadays, applying convolutional filters on signals is pretty much a standard process in signal processing studies that leverage deep learning \cite{morales2016deep, zeng2014convolutional, yang2015deep}. Convolutional neural networks (CNNs) can learn to find features in a multidimensional input while being less sensitive to the exact location of the feature in the input \cite{lecun2015deep}. In the forward pass of a convolutional layer, multiple filters are applied to the input. 
It means that in a trained neural net, multiple features can be leaned in one single convolutional layer.

Recurrent neural networks (RNN) have shown great performance in analysing sequential data such as machine translation, time-series prediction, and time-series classification \cite{cho2014learning, zhang2000predicting, wang2017time, murad2017deep, yang2015deep, Ordonez2016}. RNNs can capture long-term temporal dependencies which is quite useful for solving our problem. \cite{Che2018} For example, they might learn that ``climb'' state in a UAV auto-pilot usually follows ``take off''. Therefore, while it is outputting ``take off'' it anticipates what the next state will probably be and as soon as its input features start shifting, it detects the onset of a state change. It will help the model to better predict the system's behavior and be quicker to detect state changes in a way that could hardly be achieved with classic methods.
Therefore, in this paper, we combine the CNNs and RNNs to create what is known as a hybrid deep neural network \cite{wang2017time} to use for both CPD and state classification problems, in our context.  

\section{Hybrid Neural Network for State Inference} \label{sec:approach}
In this section, we describe our proposed deep learning approach for the black-box state inference task, in details. 

\subsection{The Model Architecture}
\begin{figure*}
    \centering
    \includegraphics[width=\textwidth]{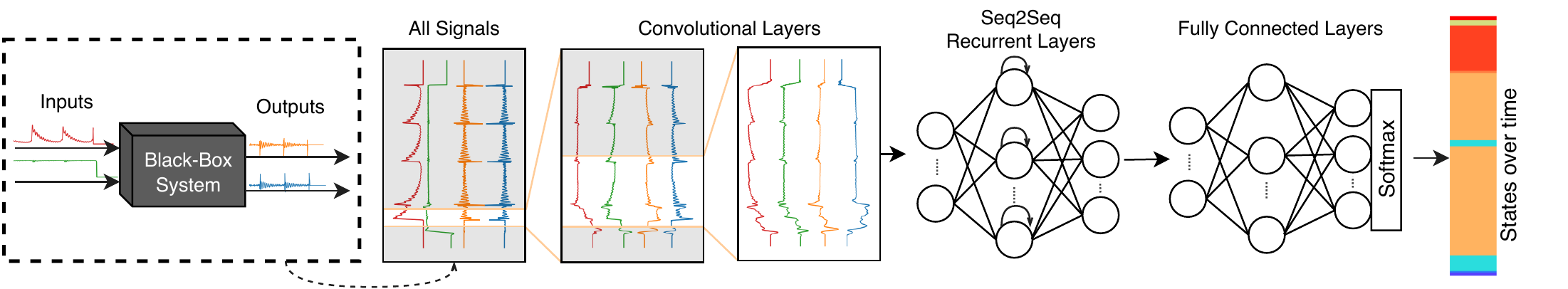}
    \caption{The input and output signals of the black-box system are captured as a multivariate time series; they are processed in a deep neural network that consists of 3 sections: convolutional, recurrent, and dense (fully connected) to predict the system's internal state and its changes over time.}
    \label{fig:general_net}
\end{figure*}

The goal of this study is to infer the states of a running software system, over time. Given that our assumption is we don't have access to the source code (or part of it), we only leverage the values of inputs and outputs of the system, over time. As can be seen in figure~\ref{fig:general_net}, we capture all the inputs and outputs of the system as a time series and then process it in a DNN. The architecture of our proposed model is a hybrid DNN which is inspired by models proposed in the field of Human Activity Recognition (HAR). This task is quite similar to the subject of our paper in the sense that they both take in a multivariate time series-data (from sensor readings) and output the state of the system that generated those readings (see section~\ref{sec:related_work_har} for more details on HAR papers). This DNN is made of three parts in sequence: 1) Convolutional, 2) Recurrent, and 3) Fully connected layers. This architecture addresses the aforementioned traditional methods' challenges; each part serves a different purpose in this process, as follows.

Convolutions, being more generalized than simple sliding windows, can discover patterns and features in the signals, both in temporal and in spatial (how signals affect each other) dimensions \cite{wang2017time}. 
The convolutional layers' flexibility allows them to learn some typical preprocessing operations. For example a moving average or a discrete derivative can be learned as simple convolutional filters. They also help the model to be more resilient to varying time delays between noticing a deviation in input signals and the reaction that will appear in the output signals. Applying convolutional layers in sequence has been shown to result in each layer learning more complex features than the previous layers \cite{zeiler2014visualizing}.
The number of layers, filters, and the kernel size are hyper-parameters that should be selected based on the size of data and the complexity of the system being modeled.
Using a sequence of convolutions with a) increasing number of filters and the same kernel size, b) same number of filters and increasing kernel size, and c) decreasing filters with increasing kernel sizes are all different approaches that have been used in the literature by well-known architectures such as VGG and U-net \cite{simonyan2014very, ronneberger2015u}. We will discuss more details of our CNN layers in Section~\ref{sec:architecture_detail}.

Convolutions are quite powerful in discovering local features. To capture long-term features, recurrent layers which learn sequences of data are leveraged. For example, in our case, they can learn that ``accelerate'' and ``take off'' states only happen in the start of the states sequence, and each ``take off'' state is usually followed by a ``climb'' state. The type of recurrent cell to use (LSTM, GRU, etc.), how many cells to unravel in the layer, and the number of layers are also hyper-parameters that need to be tuned depending on the size and complexity of system under study.

Finally, one or more dense (fully connected) layers in the end are a common way of reducing the dimensions to match expected output dimensions. 
If there are only two states, the last layer can have a sigmoid activation function and be of shape $L$ (the length of the input), otherwise, to match the one-hot encoding of labels, an output of shape $L\times N_s$ with softmax activation along the second axis ($N_s$) is required ($N_s$ being the number of possible states).

In terms of loss function to optimize in the training process, a good choice is a dice overlap loss function, which is used in image semantic segmentation tasks as well. An important property of this loss function is not getting negatively affected by class imbalances \cite{milletari2016v, sudre2017generalised}.

\subsection{Data Encoding}
The input/output values of the black-box system create a multivariate time-series ($T_k$), which can be defined as a set of $n$ univariate time series ($V_i$) of the same length $l_k$. Each $V_i$ corresponds to the recorded values for one of the inputs or outputs of the system:
\begin{equation} \label{eq:T_k}
    T_k = \{{V_1}_k, {V_2}_k, \ldots, {V_n}_k\}
\end{equation}
\begin{equation} \label{eq:l_k}
    |{V_1}_k|=|{V_2}_k|=\ldots=|{V_n}_k|=l_k 
\end{equation}

Note that, as figure~\ref{fig:general_net} shows, we take both inputs and outputs as part of the time-series data to be fed as input into our deep learning models. This is to make sure we can model state-based behavior of the system, where the current state depends not only on the inputs, but also on the last state(s) (captured as previous outputs) of the system. As an example, from our case study, if the outputs are not taken into account 
a mid-flight ``descend'' state and the ``approach'' state right before landing
are indistinguishable, using the sensor readings (inputs) alone. 

Having such a time-series, the only remaining pieces from a training set are the labels. Unlike the input/output values (the features in the data set) the labels are not usually given. Our method to infer the labels is a supervised approach. Thus, we need the domain expert to manually label each individual time stamp with a state name/ID. In practice, what they would do is to identify the approximate time that a state change happens and assign the new state to one of the previous states labels or define a new label for this new state.
Thus we encode the states information over time as a set of tuples in the form of $(t_s, s)$ where $t_s$ denotes the timestamp where the system entered state $s$. We show the set of all possible states with $S$ ($s \in S$) and define $N_s$ as the cardinality of this set. 
\begin{equation}\label{eq:change_point}\begin{split}
    CP_k {}&{}= \big\{ (t_{s_1}, s_1), (t_{s_2}, s_2), \ldots, (t_{s_l}, s_l) \big\}\:,\; s_i \in S \\
    N_s  {}&{}= |S|
\end{split}
\end{equation}
So in summary, the dataset consists of $N$ pairs of the I/O values as features and their state information as labels $\big\{(X=T_k,\;y=CP_k)|1\leq~k\leq~N\big\}$. 

\subsubsection{Data Preprocessing}
Before being fed into the model $\mathcal{F}$ (as defined below), the inputs and labels need some preprocessing. 
\begin{equation}\label{eq:model_F}
    \mathcal{F}(\delta(T), m) \colon \mathbb{R}^{L\times n}\times\mathbb{R}^L\to S^{L}.
\end{equation}
To run more efficiently, TensorFlow expects all the inputs to have the same length. To do that, the shorter $T_k$s should be zero-padded to length $L = max\{l_k\}$. The padding function $\delta$ does that.
Therefore, eventually, the input to the model will be $T_k$s that are rearranged to form a tensor of shape $n \times L$ along with a padding mask (denoted with $m$). 
The mask tells the model where the tail starts so the model can ignore all the zeros from there on. 
\begin{equation}\label{eq:model_as_function}
\begin{split}
    \hat{O} {}&{}= \langle \hat{o}_i \in S \rangle^L_{i=1} = \mathcal{F}\left(\left[ {\delta(V_1)}^\intercal \: {\delta(V_2)}^\intercal \; \ldots \; {\delta(V_n)}^\intercal \right],m\right) \\
    m {}&{}= \delta(\vec{\mathds{1}}_l) \quad\text{i.e.}\quad \langle {m}_j \rangle^l_{j=1} = 1\:, \: \langle {m}_j \rangle^L_{j=l+1} = 0
\end{split}
\end{equation}
Here $l$ denotes the length of the input before padding. It is equal to $l_k$ for the $k$th training data ($T_k$).

As defined in \eqref{eq:change_point}, $CP_k$s are tuples of $(t, s)$ which indicate the system have gone into state $s$ at time $t$. To train the model, $CP_k$ needs to be expanded into a vector of length $L$ denoted by $O$ where each element $o_t$ holds the state at time $t$. To define it formally, the elements can be derived from $CP_k$ using the following formula:
\begin{equation} \label{eq:output}
\begin{split}
O = \langle  \forall t \in \mathbb{N}_L : s_i \:|\:
           (& t_{s_i}, s_i) \in CP_k \: \land  \\
            & t_{s_i} = max\{t_{s_j} \:|\: (t_{s_j}, s_j) \in CP_k \land t_{s_j}\leq t\} \rangle
\end{split}
\end{equation}
For example: Suppose $L=10$ and $CP = \{(0, a),\:(3, b),\:(5, c),\:(8, a)\}$ 
then $O = \langle a\:a\:a\:b\:b\:c\:c\:c\:a\:a \rangle$.
If there are more than two possible states ($N_s > 2$), $O$ needs to be one-hot encoded, at this stage.

\subsection{The Model Implementation} \label{sec:architecture_detail}
The first few layers of the model are convolutional layers. We have used 5 convolutional layers with 64 filters each and a growing kernel size.
The intuition behind this design is that starting with a small kernel guides the training in a way that the first layers learn simpler more local features that fits in their window (kernel size).
Kernel sizes started with 3 since it is a common number in the literature for kernel sizes, then we used multiples of 5 from 5 to 20. 
The rationale behind choosing 5 is because the sampling frequency is 5, so each layer with a kernel size of $5n$ processes a whole $n$ seconds worth of simulation data, in each step.
Stopping at kernel size of 20 was a compromise between generalizability and model size. Generally, a larger model has more learning capacity, but it is also more prone to over-fitting. The current models are the smallest we could make the models (to avoid over-fitting), without compromising the performance.

\begin{figure*}
    \centering
    \includegraphics[width=\linewidth]{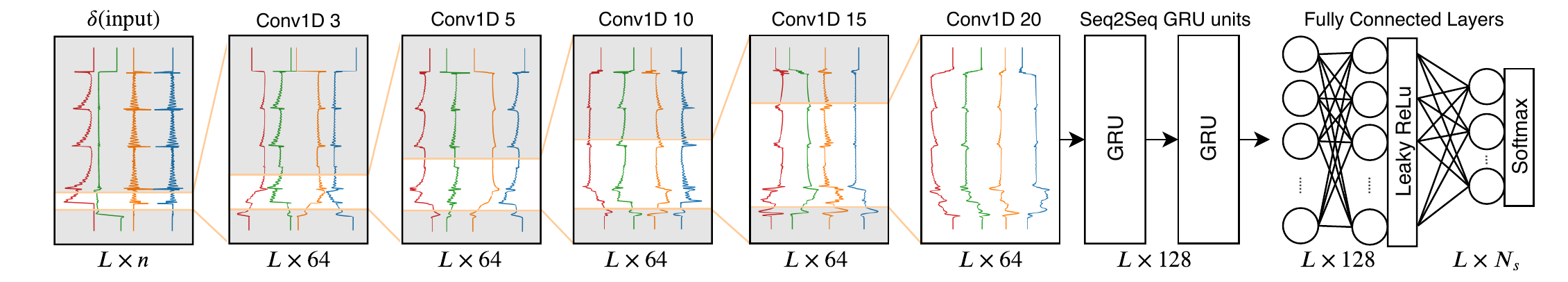}
    \caption{Model architecture in a nutshell. Tandem convolutional layers with increasing kernel size fed into two sequence-to-sequence recurrent layers with 128 GRU cells each, which is then fed into dense layers to output the predicted system state, as a list of one-hot encoded states. $\hat{O}$ will be the result of applying argmax operation on the last layer's output. $L=18000,\: N_s=25$}
    \label{fig:model_arch}
\end{figure*}

Same compromise was made in the second section of the model (Recurrent layers), the sweet spot for hyper-parameters here was to use two GRU layers with 128 cells each. Their output was fed into a fully connected layer with 128 neurons with a leaky ReLU ($\alpha=0.3$) activation function \cite{maas2013rectifier} and finally to a dense layer with $N_s=25$ units with softmax activation.
We used Adam optimizer \cite{kingma2014adam} that could converge in 60-80 epochs, i.e. validation accuracy plateaued. The full architecture can be seen in figure~\ref{fig:model_arch}.

\section{Empirical Evaluation} \label{sec:experiment}
In this section, we explain our empirical evaluation of the proposed approach through a case study.

\subsection{The Study Objectives}
The goals of this study is to evaluate our proposed method in terms of change point detection and state inference, in comparison to traditional techniques in this domain.  Therefore, our research questions are as follows:

\subsubsection{RQ 1) How does our proposed technique perform in detecting the state changes?}
The goal of this RQ is to see how close the predicted state-change times are to the real state-change times.  In other words, in RQ1, we do not predict the exact state labels and are only interested in predicting the change.
To answer this question, we compare the performance of our proposed approach with several traditional baselines (see \ref{sec:CPD_baseline}), in terms of modified precision, recall, and F1 scores that are introduced in section \ref{sec:CPD_metrics}.

\subsubsection{RQ 2) How well does our proposed technique predict the internal state of the system?}
In RQ1, we are only interested in detecting the time a state-change happens (binary classification), but here in RQ2, we extend that and are also interested in predicting the label of the new state that the system is going into (multi-class classification).  
Therefore, to answer this RQ, we change the labels from a Boolean (changed/not changed) to the actual collected labels.

Note that for both RQs, in our empirical study, to evaluate our approach, we use the source code to collect the exact time a state-change happens and the actual state labels (ground truth). However, in practice, labeling the training set is supposed to be done by the domain expert in a black-box manner. This is not an infeasible task or extra overhead. Monitoring the logs and identifying the current system state is in fact part of the developers/testers regular practice during inspection and debugging. All we provide here is a tool that given a partial labeling (only on the training set), automatically predict the state labels and the state-change times, for future flights.  Also note that even though we use the source code to label the training set, we still look at the test set as a black-box and don't leak any information.

\subsubsection{RQ 3) How much does the proposed model owe its performance to being a hybrid model?}
The proposed model architecture, inspired by the related work, combines the power of convolutional and recurrent layers. It has been shown in those contexts that using this combination is beneficial over using a fully-convolutional architecture or a recurrent architecture without any help from convolutions. To answer this question we trained two other models one without the convolutional part and one without the GRU layers. We compare these two with the proposed model.

\subsection{Evaluation Metrics}
\subsubsection{RQ1 (CPD) Performance Metrics} \label{sec:CPD_metrics}
Given that in RQ1 there is an inherent class imbalance (there are far more points where a change has \textit{not} happened compared to points with a state-change positive label), we avoid using accuracy and report both precision and recall. 
However, the original precision/recall metrics require some modifications due to the difficulty of predicting the exact time stamp that a state-change happened. To handle this, similar to related work \cite{Truong2018ChangePointSurvey}, we use a tolerance margin $\tau$. If a detected state-change ($\in\hat{CP}_k$) is within $\pm\tau$ of a true change ($\in{CP}_k$) , we call the prediction a True Positive, otherwise it is a False Positive. Similar adjustment to definition is applied for True Negative and False Negative. 
Formally speaking, we define predicted change points for $k$-th sample as:
\begin{equation} \label{eq:cp_hat}
\hat{CP}_k = \big\{(t, \hat{o}_t)\: |\: \hat{o}_t \neq \hat{o}_{t-1} \big\}
\end{equation}
Please note that in \eqref{eq:cp_hat}, $\hat{o}_t$ refers to $t$-th element of output vector $\hat{O}$, as previously defined in \eqref{eq:model_as_function}. Based on that the confusion matrix elements are calculated as:
\begin{equation} \label{eq:metrics}
\begin{split}
TP ={}&{}\Big|\big\{ (\hat{t}, \hat{s}_t) \in \hat{CP}_k \;\big|\; \exists\: (t, s_t) \in CP_k \;\text{s.t.}\; |t - \hat{t}| < \tau\big\}\Big| \\
FP ={}&{}\Big|\big\{ (\hat{t}, \hat{s}_t) \in \hat{CP}_k \;\big|\; \nexists\: (t, s_t) \in CP_k \;\text{s.t.}\; |t - \hat{t}| < \tau\big\}\Big| \\
FN ={}&{}\Big|\big\{ (t, s_t) \in CP_k \;\big|\; \nexists\: (\hat{t}, \hat{s}_t) \in \hat{CP}_k \;\text{s.t.}\; |t - \hat{t}| < \tau\big\}\Big| 
\end{split}
\end{equation}
With these in mind, we measure precision, recall, and their harmonic mean F1 Score with three values for $\tau$: 1, 3, and 5 seconds. The smaller the tolerance is the stricter the definitions become and the lower the numbers are. 

\subsubsection{RQ2 (State detection) metrics}
In RQ2, we have a multi-class classification problem and thus multiple precisions/recalls will be calculated, one per class (state label). We then report the mean value across all classes. 
\begin{equation}
\begin{split}
P_s ={}&{}\big\{\hat{s}_t \in \hat{O}_k \;\big|\; \hat{s}_t = s\big\} \\
T_s ={}&{}\big\{s_t \in O_k \;\big|\; s_t = s\big\} \\
TP_s ={}&{}\big\{\hat{s}_t \in P_s \;\big|\; \hat{s}_t = s_t \in O_k\big\} \\
\end{split}
\end{equation}
$$Precision =\frac{1}{N_s}\sum_{s=1}^{N_s} \frac{|TP_s|}{|P_s|} \quad,\quad Recall = \frac{1}{N_s}\sum_{s=1}^{N_s} \frac{|TP_s|}{|T_s|}$$

\subsubsection{RQ3}
We used the same metrics as RQ1 and RQ2.

\subsection{Comparison Baselines} 
\subsubsection{RQ1 (CPD) baselines} \label{sec:CPD_baseline}
We used `ruptures' library developed by authors of a recent CPD survey study \cite{Truong2018ChangePointSurvey}. It provides a modular framework for applying several CPD algorithms to univariate and multivariate data. 
As mentioned earlier two main elements of a CPD algorithm in their survey are the search method and the cost function.

We used Pelt \cite{killick2012optimal} as the most efficient exact search method. As examples of approximate search methods, we applied bottom-up segmentation and window-based methods using a default window size of 100 \cite{keogh2001online}.
However, after trying to run Pelt algorithm, we realized that it takes prohibitively longer to run compared to the approximate methods without providing much better results, so we only use the bottom-up and the window-based segmentation methods, as our CPD baselines.

For the cost function, we tried ``Least Absolute Deviation'', ``Least Squared Deviation'', ``Gaussian Process Change'', ``Kernelized Mean Change'', ``Linear Model Change'', ``Rank-based Cost Function'', and ``Auto-regressive model change'' as defined in the library. Their parameters were left as default.
To optimize the number of change points a penalty value (linearly proportionate to the number of detected change points) is added to the cost function, which limits the number of detected change points, the higher the penalty the fewer reported change points. We tried three different ratios (100, 500, and 1000) for the penalty.

\subsubsection{RQ2 (Multi-class classification) baselines}
We used a sliding window of width $w$ over the 10 time-series values and then flattened it to make a vector of size $10w$ as the features. For the labels, we used one-hot encoded state of the system.
The window sizes were chosen as same as the sizes of convolutional layers' kernel sizes (3, 5, 10, 15, 20), to make the baselines better comparable with our method. 
We used Scikit-learn's implementation of the classification algorithms: A ridge classifier (Logistic regression with L2 regularization) and three decision trees. The ridge classifier was configured to use the built-in cross validation to automatically chose the best regularization hyper-parameter $\alpha$ in the range of $10^{-6}$ to $10^6$. Each decision tree was regularized by setting ``maximum number of features'' and ``maximum depth''. For ``maximum number of features'' we tried no limits, $\sqrt{10w}$, and $\log_2{10w}$. To find best ``maximum depth'' we first tried having no upper bound and observed how deep the tree grows; then we tried multiple numbers less than the maximum, until a drop in performance was observed.

\subsubsection{RQ3 Hybrid vs. Homogeneous baselines}
We compared two versions of the model, one fully convolutional and one fully recurrent, with the full hybrid model to see if combining RNNs and CNNs has an added value or the same results could be achieved using only one type of layers.

\subsection{Dataset and the Data Collection Process}
\begin{figure}
    \centering
    \includegraphics[width=\columnwidth]{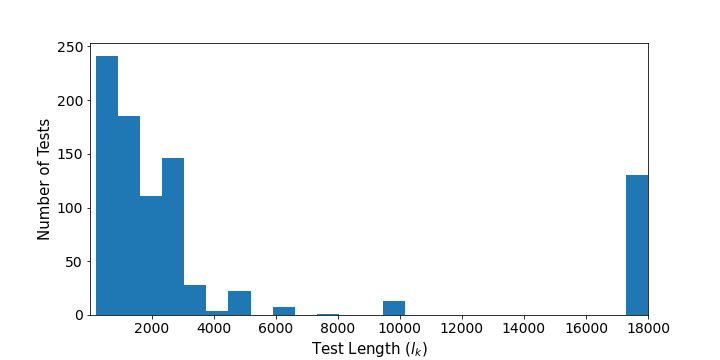}
    \Description[Number of tests histogram]{A histogram chart maxing on numbers less than 2,000 and over 17,500}
    \caption{Distribution of flight log lengths for the $N=888$ logs (out of the original 948 available logs) which were kept in the dataset ($200 \leq l_k \leq 20,000$)}
    \label{fig:test_lengths}
\end{figure}

We ran 948 existing test cases from MicroPilot's test repository using a software simulator\footnote{It is developed by MicroPilot Inc and provides an accurate simulation of the aerodynamic forces on the aircraft, the physical environment irregularities (e.g. unexpected wind gusts), and noises in sensor readings} and collected the logged flight data, over time. The test cases are system-level tests. Each test case includes a flight scenario for various supported aircraft. A flight scenario goes through different phases in a flight such as ``take off'', ``climb'', ``cruise'', ``hitting way points'', and ``landing''.
We sampled input and output values (listed in Table~\ref{tab:in_outs}) at 5 Hz rate, which is the rate that AutoPilot reads the sensor values and performs the calculations required to update its output values at.
\begin{table}
    \caption{The $n=10$ collected I/Os of AutoPilot. The inputs are sensor readings and the outputs are the servo position update commands. All these I/Os over time are used as the inputs of the state prediction model.}
    \label{tab:in_outs}
    \centering
\begin{tabularx}{\columnwidth}{lX}
                                                                                                                    \toprule
\multicolumn{2}{l}{\textbf{Inputs}}                                                                              \\ \midrule
Pitch     & The angle that aircraft's nose makes with the horizon around lateral axis                            \\ 
Roll      & The angle of aircraft's wings make with the horizon around longitudinal axis                         \\ 
Yaw       & The rotation angle of aircraft around the vertical axis                                              \\ 
Altitude  & AGL\footnotemark Altitude                                                                            \\ 
Air speed & Speed of the aircraft relative to the air                                                            \\ \midrule
\multicolumn{2}{l}{\textbf{Outputs}}                                                                             \\ \midrule
Elevator  & Control surfaces that control the Pitch                                                              \\ 
Aileron   & Control surfaces that control the Roll                                                               \\ 
Rudder    & Control surface that controls the Yaw                                                                \\ 
Throttle  & Controller of engine's power, ranges from 0 to 1                                                     \\ 
Flaps     & Surfaces of back of the wings that provide extra lift at low speeds, usually used during the landing \\ \bottomrule
\end{tabularx}
\end{table}
\footnotetext{Above Ground Level}
Out of the 948 flight logs, we omitted 60 that were either too short or too long (shorter than 200 samples or longer than 20k samples). Figure~\ref{fig:test_lengths} shows the distribution of the remaining log lengths. The maximum length ($L$) was 18,000 samples.

The dataset was randomly split into three chunks of 90\%, 5\%, and 5\% for training, validation, and testing, where each sample corresponds to one test execution. Note that separate test and validation sets are needed to facilitate proper hyper-parameters tuning, without leaking information. 


\subsection{Experiment Execution Environment} \label{sec:machines_config}
Training and evaluation of the deep learning model was done on a single node running Ubuntu 18.04 LTS (Linux 5.3.0) equipped with Intel Core i7-9700 CPU, 32 gigabytes of main memory, and 8 gigabytes of GPU memory on a NVIDIA GeForce RTX 2080 graphics card.
The code was implemented using keras on TensorFlow 2.0.

The baseline models could not fit on that machine, so two nodes on Compute Canada's Beluga cluster, one with 6 CPUs and 75GiB of memory and one with 16 CPUs and 64GiB of memory, were used to train and evaluate them.

\subsection{Results} \label{sec:results}
In this section, we present the results of the experiments and answer the two research questions. 
\subsubsection{RQ1 Results: CPD Performance}
\begin{table*}
\caption{Change point detection precision, recall, and F1-score calculated for the baseline methods using three values of tolerance ($\tau$) for multiple configurations.}
\label{tab:rq1-results}
\resizebox{\textwidth}{!}{%
\begin{tabular}{llcccccccccc}
\toprule
  \textbf{Cost Function} &
  \textbf{Search Method} &
  \textbf{Penalty} &
  \textbf{Prec.} &
  \textbf{Recall} &
  \textbf{F1} &
  \textbf{Prec.} &
  \textbf{Recall} &
  \textbf{F1} &
  \textbf{Prec.} &
  \textbf{Recall} &
  \textbf{F1} \\
                                                  &              &   & \multicolumn{3}{c}{$\tau=1$s} & \multicolumn{3}{c}{$\tau=3$s} & \multicolumn{3}{c}{$\tau=5$s} \\  \toprule
\multirow{2}{*}{\textbf{Autoregressive Model}}     & Bottom Up    & 1000 & 10.43\%    & 75.44\%    & 18.33\%   & 21.21\%    & 80.32\%    & 33.55\%   & 28.94\%    & 81.22\%    & 42.68\%   \\
                                                   & Window Based & 100  & 2.94\%     & 3.98\%     & 3.38\%    & 8.53\%     & 11.41\%    & 9.76\%    & 12.89\%    & 17.54\%    & 14.86\%   \\ \midrule
\multirow{2}{*}{\textbf{Least Absolute Deviation}} & Bottom Up    & 500  & 7.32\%     & 52.54\%    & 12.85\%   & 17.52\%    & 87.73\%    & 29.20\%   & 25.02\%    & 88.95\%    & 39.05\%   \\
                                                   & Window Based & 500  & 5.24\%     & 8.31\%     & 6.42\%    & 15.20\%    & 24.03\%    & 18.62\%   & 21.79\%    & 38.25\%    & 27.76\%   \\ \midrule
\multirow{2}{*}{\textbf{Least Squared Deviation}}  & Bottom Up    & 1000 & 7.44\%     & 85.09\%    & 13.68\%   & 16.40\%    & 89.81\%    & 27.74\%   & 24.16\%    & 90.47\%    & 38.14\%   \\
                                                   & Window Based & 500  & 3.59\%     & 6.79\%     & 4.70\%    & 10.27\%    & 16.51\%    & 12.66\%   & 16.18\%    & 26.84\%    & 20.19\%   \\ \midrule
\multirow{2}{*}{\textbf{Linear Model Change}}      & Bottom Up    & 100  & \textbf{37.59\%}    & 28.98\%    & \textbf{32.73\%}   & \textbf{45.20\%}    & 38.39\%    & \textbf{41.52\%}   & \textbf{48.07\%}    & 41.36\%    & \textbf{44.46\% }  \\
                                                   & Window Based & 500  & 6.70\%     & 4.14\%     & 5.12\%    & 20.50\%    & 13.05\%    & 15.95\%   & 38.78\%    & 26.77\%    & 31.67\%   \\ \midrule
\multirow{2}{*}{\textbf{Gaussian Process Change}}  & Bottom Up    & 100  & 3.77\%     & \textbf{92.23\%}    & 7.25\%    & 8.99\%     & \textbf{92.23\%}    & 16.39\%   & 13.53\%    & \textbf{92.23\%}    & 23.60\%   \\
                                                   & Window Based & 100  & 2.94\%     & 3.95\%     & 3.37\%    & 8.69\%     & 11.50\%    & 9.90\%    & 13.64\%    & 18.30\%    & 15.63\%   \\ \midrule
\multirow{2}{*}{\textbf{Rank-based Cost Function}} & Bottom Up    & 100  & 13.45\%    & 60.19\%    & 21.98\%   & 19.49\%    & 80.10\%    & 31.35\%   & 22.98\%    & 87.23\%    & 36.38\%   \\
                                                   & Window Based & 100  & 8.10\%     & 13.70\%    & 10.18\%   & 15.72\%    & 30.73\%    & 20.80\%   & 21.38\%    & 46.64\%    & 29.32\%   \\ \midrule
\multirow{2}{*}{\textbf{Kernelized Mean Change}}   & Bottom Up    & 100  & 4.13\%     & 3.24\%     & 3.63\%    & 12.22\%    & 8.14\%     & 9.77\%    & 15.38\%    & 10.58\%    & 12.54\%   \\
                                                   & Window Based & 100  & 2.82\%     & 3.00\%     & 2.91\%    & 10.14\%    & 8.40\%     & 9.19\%    & 13.64\%    & 12.61\%    & 13.10\%   \\ \bottomrule

\end{tabular}%
}
\end{table*}

Table~\ref{tab:rq1-results} shows the results of running CPD algorithms for various configurations (as described in \ref{sec:CPD_baseline}). For each search method and cost function pair only one of the penalty values which resulted in the highest F1 scores for all $\tau$ values is reported. 

The first observation from the results is that as values of $\tau$ increases the scores get better. This was expected, since larger values relax the constraints on which detected change points are considered as a true positive. Another observation is that the bottom-up segmentation consistently outperforms the window-based segmentation method. We can also see that the linear cost function beats all the other ones, in terms of precision. The Gaussian cost function achieves much higher recall values costing it a huge loss in precision. It means this cost function results in detecting numerous change points spread across the time axis, so there is a good chance of having at least one change point predicted close to each true change point (hence the high recall), but also there are a lot of false positives, which leads to a low precision. 

\begin{table}
\caption{Change point detection precision, recall, and F1-score calculated, on the test data, for our proposed model, using three values of tolerance ($\tau$) compared with the respective $\tau$'s best F1 score among baseline methods}
\label{tab:rq1-2-results}
\begin{tabularx}{\columnwidth}{cXXXX}
\toprule
  $\mathbf{\tau}$ &
  \textbf{Prec.} &
  \textbf{Recall} &
  \textbf{F1 score} & 
  \textbf{Baseline F1}  \\ \midrule
1s & 56.77\% &	79.32\% &	66.18\% & 32.73\% \\
3s & 69.58\% &	88.88\% &	78.06\% & 41.52\% \\
5s & 79.82\% &	91.87\%	&   85.42\% & 44.46\% \\\bottomrule
\end{tabularx}
\end{table}

Measuring the same metrics on how our model performs on the test data shows better scores, almost twice the F1 score of the best performing baseline (see Table~\ref{tab:rq1-2-results}). Please note that unlike machine learning algorithms (such as ours), CPD algorithms do not have a separate training and testing phases. This fact works in their favor (by using the entire dataset for prediction and not just the training set), but still our model outperforms them.

In terms of execution cost, running all 42 different settings of CPD algorithms on the whole dataset took a bit over 12 hours in the cloud using 16 CPUs and 64GB of main memory. The deep learning model on the other hand takes about an hour to train (which only needs to be done once), on a smaller machine (see section~\ref{sec:machines_config}). It made predictions on the whole dataset in less than a minute.
So to answer RQ1, our method has shown $(66.18/32.73)-1=102.20\%$ improvement in F1 score with $\tau=1s$, $(78.06/41.52)-1=88.00\%$ with $\tau=3s$, and $(85.42/44.46)-1=92.13\%$ with $\tau=5s$; almost doubling the score compared to the baselines. 

\begin{rqanswer}
Our model, which requires less memory compared to traditional CPD algorithms, improved their best performance by up to 102\%, measured by F1 score, in less execution time.
\end{rqanswer}

\subsubsection{RQ2 Results: Multi-class Classification Performance}
To answer RQ2, we first compare different configurations of the baseline methods using the F1 score (harmonic mean of precision and recall) on the test data. The results are presented in Table~\ref{tab:rq2-1-results}.

\begin{table}
\caption{Precision, recall, and F1 score of ridge classifiers (linear classifiers with L2 regularization) and decision tree classifiers (DT) with different sliding window widths ($w$). For each algorithm on each $w$ several hyper-parameters were applied producing 152 different models. In this table, we only show the results of the best performing model in each group.}
\label{tab:rq2-1-results}
\resizebox{\linewidth}{!}{%
\begin{tabular}{clcclll}
\toprule
\textbf{w} &
  \multicolumn{1}{c}{\textbf{Classifier}} &
  \multicolumn{1}{c}{\textbf{\begin{tabular}[c]{@{}c@{}}Max\\ Depth\end{tabular}}} &
  \multicolumn{1}{c}{\textbf{\begin{tabular}[c]{@{}c@{}}Max\\ Features\end{tabular}}} &
  \multicolumn{1}{c}{\textbf{Prec.}} &
  \multicolumn{1}{c}{\textbf{Recall}} &
  \multicolumn{1}{c}{\textbf{F1}} \\ \midrule
3  & Ridge & -   & -          & 71.39\% & 20.73\% & 32.13\% \\
3  & DT    & -   & -          & 69.21\% & 82.36\% & 75.21\% \\ \midrule
5  & Ridge & -   & -          & 69.15\% & 21.89\% & 33.26\% \\
5  & DT    & 100 & -          & 68.37\% & \textbf{83.16\%} & 75.04\% \\ \midrule
10 & Ridge & -   & -          & 71.97\% & 24.02\% & 36.02\% \\
10 & DT    & 260 & -          & 67.94\% & 79.14\% & 73.12\% \\ \midrule
15 & Ridge & -   & -          & 76.87 & 25.90\% & 38.75\% \\
15 & DT    & -   & $\sqrt{10w}$ & 69.06\% & 80.76\% & 74.45\% \\ \midrule
20 & Ridge & -   & -          & \textbf{80.38\%} & 26.50\% & 39.86\% \\
20 & DT    & 175 & $\sqrt{10w}$ & 73.21\% & 82.16\% & \textbf{77.42\%} \\  \midrule
   & \multicolumn{3}{l}{Our Proposed Method} & \textbf{86.29\%} & \textbf{95.04\%} & \textbf{90.45\%} \\
\bottomrule
\end{tabular}%
}
\end{table}

Comparing the baseline methods with our proposed method (the last row) in Table~\ref{tab:rq2-1-results} shows that our model outperforms all baselines. Comparing it with the model with the best F1-score shows a $(86.29/73.21)-1=17.87\%$ improvement in precision as well as a $(95.04/82.16)-1=15.68\%$ improvement in recall that means $(90.45/77.42)-1=16.83\%$ overall improvement in F1-score. 

\begin{figure*}
    \centering
    \includegraphics[width=\linewidth]{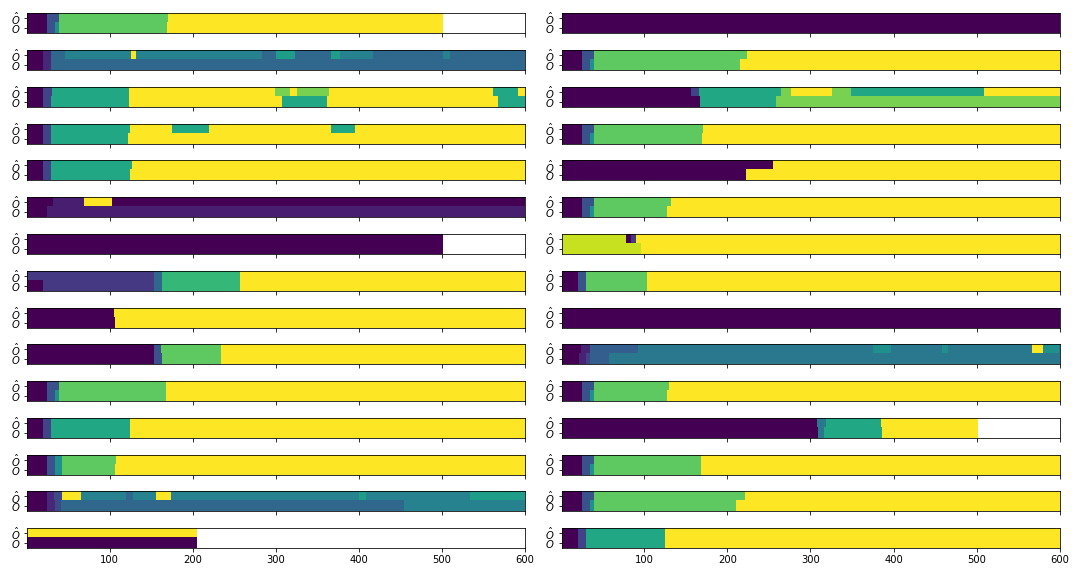}
    \Description[States in time diagram]{A figure that color-codes the true and predicted states}
    \caption{Evaluation of the model on 30 random test data. Each graph shows the states in one run of the system. The colors show the states. The top-half of each plot depicts model's prediction of the system states ($\hat{O}$) and the bottom-half shows the true labels($O$). Since the output is one-hot encoded, the item with the most probability is used as the predicted label at each point in time. X-axis is the time axis. Only the first 600 samples (2 minute of simulation) are shown to improve legibility.}
    \label{fig:test_0}
\end{figure*}

To have a feeling of how good our predictions are in practice, Figure~\ref{fig:test_0} shows the output of our model side by side with the ground truth. The horizontal axis shows sample ID (time) and the states are color coded. As it is seen, our algorithm performs better when the state changes are farther apart. Also there are some state changes that happen quite briefly which are not detected. That is not to a great surprise since it takes some time for state changes to be reflected in the outputs and those might not have got any chance.

The classical models only see one window of the data at a time, convolutional layers on the other hand are more generalized and flexible since each filter in each layer is comparable to a sliding window. As we saw in Table~\ref{tab:rq2-1-results}, a larger window size means a higher performance. However, it gets significantly more difficult to train a model with large window sizes. In addition, convolutions can automatically learn preprocessing steps that could be beneficial such as a moving average. Each convolutional filter can learn a linear combination of its inputs. So when the convolutional layers are stacked on each other, with non-linear activation functions in between, the hypothesis space they can learn becomes quite large, probably much larger than most of the classical ML algorithms here. Also, they are still quite efficient (more efficient than baselines) due to parameter sharing and their high parallelizability.

The fact that the performance improves as the window size increases indicates the positive effect of being able to see longer-term relations in detecting the system's state. Recurrent cells (such as GRU) can capture long-term dependencies (that do not necessarily fall into one window) and learn sequences. This is one of the major differences between an RNN model and others, such as decision trees, which do not have such a notion of a ``long-term memory'' as LSTM/GRU neural networks do. All a decision tree could see is the values in a sliding window.

In terms of the training complexity (time and memory), our method is superior as well. That can largely be attributed to the use of deep learning. In baseline models, as the window size $w$ grows the training and evaluation complexity grows, up to a point that they ran out of memory -- consuming all the 47GB of main memory and swap area. This forced us to train them in the cloud. Meanwhile, as mentioned earlier, the deep learning model could be trained on a 8GB GPU in roughly an hour. (see section \ref{sec:machines_config} for the machines' specs). Also, the decision tree training was not parallelized using only one core of the CPU, while virtually all deep learning models can be heavily parallelized on a GPU/TPU. 

\begin{rqanswer}
Our model, which requires less than half as many CPUs and 70\% as much memory compared to the best performing classical ML model, improved their best performance by up to 17\%, measured by F1 score, in less execution time.
\end{rqanswer}

\subsubsection{RQ3 Results: Hybrid vs. Homogeneous model}
As the results in table~\ref{tab:rq3} suggest, using a hybrid architecture in this problem delivers more than sum of its parts, outperforming the fully convolutional and fully recurrent baselines. 
We can also see how the RNN baseline got closer results to the full model, suggesting the important role it plays in capturing long-term relations in the data and inferring the system's internal state.

You might also notice that the results in the last column differ a little (around 1\% in absolute value) from their corresponding results in tables~\ref{tab:rq2-1-results} and \ref{tab:rq1-2-results}. That is due to randomizations in splitting the data into training, testing, and validation sets.
\begin{table}
    \centering
    \caption{Comparing the hybrid model's performance in both regards with the the its homogeneous sub-models.}
    \label{tab:rq3}
    \begin{tabular}{llccc}
        \toprule
        {} & $\tau$ &  \textbf{RNN only} &  \textbf{CNN only} &  \textbf{Full Model} \\
        \midrule
        Precision & 1s &         45.31\% &    38.12\% & 53.84\% \\
        Recall & 1s    &         60.56\% &    58.50\% & 67.94\% \\
        F1 & 1s        &         51.84\% &    46.16\% & 60.06\% \\ \midrule
        Precision & 3s &         56.06\% &    50.97\% & 72.00\% \\
        Recall & 3s    &         78.00\% &    69.12\% & 88.56\% \\
        F1 & 3s        &         65.25\% &    58.69\% & 79.44\% \\ \midrule
        Precision & 5s &         68.75\% &    67.38\% & 78.56\% \\
        Recall & 5s    &         81.81\% &    73.75\% & 93.75\% \\
        F1 & 5s        &         74.69\% &    70.44\% & 85.50\% \\ \midrule
\multicolumn{2}{l}{Classification Prec.} &    81.56\% & 71.56\% & 88.44\% \\
\multicolumn{2}{l}{Classification Recall}&    91.88\% & 86.88\% & 94.50\% \\
\multicolumn{2}{l}{Classification F1}    &    86.44\% & 78.44\% & 91.38\% \\
        \bottomrule
        \end{tabular}
\end{table}

\begin{rqanswer}
The hybrid architecture performs better than a comparable RNN model or fully convolutional model, however the recurrent section plays a more important role in the model's performance.
\end{rqanswer}

\subsection{Limitations and Threats to Validity} \label{sec:threats_to_validity}
One of the limitations of our approach is that it might miss an input-output invariant correlation. It can happen when the input remains constant or it changes too little to reveal its relation with certain outputs, therefore remaining unobserved. However this is a shared shortcoming of dynamic analysis approaches.
We assume that during the data collection, sampling happens in regular intervals; our approach probably will have a hard time achieving high performances, working on unevenly spaced time-series data.

In terms of construct validity, we are using standard metrics to evaluate the results. However, the use of tolerance margin should be taken with caution since it is a domain-dependant variable and can change the final results. To alleviate this threats, we have used multiple margins and reported all results. 
In terms of internal validity threats, we reduced the threat by not implementing the CPD baselines by ourselves and rather reusing existing libraries. 
In terms of conclusion validity threats, we have used many (888) real test cases from MicroPilot's test repository and provided a proper train-validation-test split for training, tuning, and evaluation. 
Finally, in terms of external validity threats, our study suffers from being limited to only one case study. 
However, the study is a large-scale real-world study with many test cases. 
We plan to extend this work with more case studies from other domains, to increase its generalizability.

\section{Related Work} \label{sec:related_work}

\subsection{Time Series Change Point Detection}
Change point detection is a well-studied subject due to its wide range of applications \cite{basseville1993detection}.
Several statistical and algorithmic methods have been tried to tackle several variations of this problem \cite{chen2011parametric, hasan2014information, hsu1982bayesian, lee2017implicit, oh2002analyzing, ramos2016anomalies, chowdhury2012bayesian, reeves2007review, rosenfield2010change, wang2011non, xie2013sequential, yamanishi2004line, Lavielle1999}. 
The models vary based on: whether the whole data is available at once (offline) or it is being generated on the go (online), whether there are statistical assumptions about the data distribution \cite{takeuchi2006unifying, ide2007change}, whether the number of change points is known \cite{Truong2018ChangePointSurvey}, or whether we are dealing with a univariate or a multivariate time series, etc. 

Ives and Dakos utilized locally linear models and used statistical significance test to determine at which point the changes in model parameters are large enough to signal a change in the state \cite{Ives2012}. Blythe et al., used subspace analysis to reduce data dimensionality to keep the most non-stationary dimensions. This process helps detecting change points more effectively \cite{Blythe2012}.
Several techniques have used penalty functions to find models that best fit each segment of the signal \cite{Lavielle1999, lavielle2005using, keshavarz2018optimal, pein2017heterogeneous, khan2019deep}. 
Desobry et al., and Hido et al., proposed methods to indirectly use classifiers such as SVM to detect change points \cite{desobry2005online, hido2008unsupervised, Khan2019thesis}. We applied their approach on our data in early stages of the research but it could not perform as others.
Lee et al., trained deep auto encoder networks that learns latent features in the data to detect change points \cite{Lee2018TimeSeriesSegmentation}. Ebrahimzadeh et al., proposed what they call a pyramid recurrent neural network architecture, which is resilient to missing to detect patterns that are warped in time \cite{Ebrahimzadeh2019}.

There are also a family of methods based on Bayesian models that focus on finding changes in parameters of underlying distributions of the data \cite{Lee2018TimeSeriesSegmentation, adams2007bayesian, bai1997estimation, barry1993bayesian, erdman2008fast, ray2002bayesian}. 

Making assumptions about the data such as its distribution or the distribution of change points across the time and relying on basic statistical properties are the two major short comings of traditional CPD methods \cite{Lee2018TimeSeriesSegmentation}, which our proposed approach has overcome.

\subsection{State Model Inference}
Roughly speaking, dynamic EFSM\footnote{Extended Finite State Machines, are special kind of state machines that have conditional expressions called ``transition guards'' on their transitions \cite{lorenzoli2008automatic}. A state transition can only happen if the transition guard evaluates as true.} inference algorithms generally take a trace of ``events'' (along with perhaps some variable values) as their input \cite{walkinshaw2016inferring} to infer a generalized finite state machine. They use the events to find the state transitions and the values for detecting invariants and generating the guard conditions on the transitions. 
k Tails, Gk Tail, EDSM, and MINT are examples of these algorithms, each improving upon the previous one \cite{biermann1972synthesis, lorenzoli2008automatic, lang1998results, walkinshaw2016inferring}.  

Walkinshaw et al., proposed an algorithm and developed a tool for state model inference \cite{walkinshaw2016inferring}. Their work is based on previous endeavors on state merging algorithms such as gk-tail and k-tails \cite{lorenzoli2008automatic, biermann1972synthesis}. These methods require an execution trace of the program consisting of a list of ``events'' that occurred during program execution, such as function calls, system calls, transmitted network data, etc. Krka et al., performed an empirical study on 4 different categories of model inference algorithms to figure out what makes each group of methods more effective \cite{krka2014automatic}. Beschastnikh et al., proposed a method to mine invariants from partially ordered logs from concurrent/distributed systems \cite{beschastnikh2011mining}. Invariants can be used to augment state models \cite{beschastnikh2014inferring, beschastnikh2011leveraging}. Groz et al., use machine learning to heuristically infer state machine models of a un-resettable black-box system \cite{groz2018revisiting}, however a significant difference between our method and theirs is that their method still relies on discrete events (such as HTTP request and responses) while our method does not assume that the input and outputs contain any kind of ``events'' happening at certain times. Our method aims to search for such events as change points in a continuous stream of data as time series. 

\subsection{Using Deep Learning on Time Series Data} \label{sec:related_work_har}
Human activity recognition (HAR) is a well researched task which is quite relevant to the problem of black-box model inference. In HAR, just like in our context, a multivariate time series data is created from various sensors on a human body. The goal is to figure our what was the activity that human was performing in different time intervals. The sensors can be body worn accelerometers, or more generic sensors such as the ones found in a smart watch or a smartphone. 
Murad et al., \cite{murad2017deep} have shown deep RNNs outperform fully convolutional networks and deep belief networks in HAR task.
Hybrid models are the combination of some deep architectures \cite{wang2019deep}, such as a CNN + RNN or a CNN + a fully connected net. Morales et al., have shown the former preforms better than the latter in HAR \cite{morales2016deep}. Yao et al., \cite{deepsense} introduced a CNN + RNN architecture that outperforms the state of the art both in classification and in regression tasks. Similar results have been shown in other works such as \cite{Ordonez2016, singh2017transforming, zheng2016exploiting}, as well.

Another related topic here is the time series classification. However, time series classification techniques often output only one label classifying entire data, thus not applicable in our context. What is more related to our problem is called ``segmentation'', using the computer vision terminology (not be confused with time series segmentation, such as \cite{lemire2007better}). U-net is one of the promising auto-encoder architectures for image segmentation \cite{ronneberger2015u}. Perslev et al., developed a similar idea for time-series to capture long-term dependencies and called it U-time \cite{perslev2019u}. It is fully convolutional and does not use memory cells (recurrent cells). A fully convolutional model can perform very well, since convolutions operate locally and image segments are large chunks of pixels in the 2D space and capturing local features using neighbouring pixels is quite useful. However, it cannot necessarily be as powerful on a more limited 1D data of time-series with different characteristics from an image.  This study's design is optimized for the task of sleep phase detection, which does not have very clear boundaries between states and also the state changes are not very frequent. Therefore, the same method does not necessarily generalize to tasks such as ours, where we cannot make assumptions about frequency of state changes.

\section{Summary and Future Work} \label{sec:summary} 
In this paper, we introduced a hybrid CNN-RNN model that can be used for both CPD and state classification problems in multivariate time series. The proposed approach can be used as a black-box state model inference for variety of use cases such as testing, debugging, and anomaly detection in control software systems, where there are several input signals that control output states. We have evaluated our approach on a case study of a UAV auto-pilot software from our industry partner with 888 test cases and showed significant improvement in both change point detection and state classification. In the future, we are planning to extend this research with more case studies from open source auto-pilots. In addition, better tuning of hyper-parameters will be explored. Finally, we plan to examine the use of transfer learning to reduce the labeling overhead.

\begin{acks}
We would like to thank the anonymous reviewers for their constructive comments.
We acknowledge the support of the \grantsponsor{NSERC}{Natural Sciences and Engineering Research Council of Canada (NSERC)}, [funding reference number \grantnum{NSERC}{CRDPJ/515254-2017}].
\end{acks}

\bibliographystyle{ACM-Reference-Format}
\bibliography{main}

\end{document}